\documentclass[manuscript]{acmart}
\title\title
\usepackage{booktabs}
\usepackage{amsmath}
\usepackage{array}
\usepackage{graphicx}
\usepackage{url}
\usepackage{enumitem}
\usepackage{multicol} % Two columns *inside* a prompt-template figure only
\usepackage{float}
\usepackage{parskip}
\usepackage{tikz}
\usepackage{pgfplots}
\pgfplotsset{compat=1.18}

\newcommand{\imwutimage}[2]{%
  \IfFileExists{#1}{%
    \includegraphics[width=#2\linewidth]{#1}%
  }{%
    \fbox{\parbox[c][42mm][c]{0.85\linewidth}{\centering\small
      Missing original figure: \texttt{\detokenize{#1}}\\
      Upload this image from the original Overleaf project.}}%
  }%
}

\usepackage[most]{tcolorbox}
\tcbuselibrary{breakable}
\newtcolorbox{promptbox}[1]{
  title={#1},
  breakable,
  colback=gray!3,
  
  colframe=gray!60,
  left=1mm,right=1mm,top=0.5mm,bottom=0.5mm,
  fontupper=\ttfamily\footnotesize,
  parbox=false,
  before upper={\setlength{\parindent}{0pt}}
}

\acmJournal{IMWUT}
\setcopyright{none}
\title[ThermoLLM: LLM-Based Multi-Zone HVAC Control with a Spatiotemporal Context]{ThermoLLM: LLM-Based Multi-Zone HVAC Control with a Spatiotemporal Context}

\author{Kirtan Bhatt}
\affiliation{
  \institution{UNSW Sydney}
  \country{Australia}
}
\email{kirtan.bhatt@unsw.edu.au}

\author{Dawn Lin}
\affiliation{
  \institution{UNSW Sydney}
  \country{Australia}
}
\email{dawn.lin@student.unsw.edu.au}

\author{Matthew Amos}
\affiliation{
  \institution{CSIRO}
  \city{Newcastle}
  \country{Australia}
}
\email{matt.amos@csiro.au}

\author{Flora Salim}
\affiliation{
  \institution{UNSW Sydney}
  \country{Australia}
}
\email{flora.salim@unsw.edu.au}

\author{Wen Hu}
\affiliation{
  \institution{UNSW Sydney}
  \country{Australia}
}
\email{wen.hu@unsw.edu.au}

\begin{abstract}
\textbf{ABSTRACT}\par\smallskip
Multi-zone HVAC control is a spatial decision problem in which indoor thermal evolution and control decisions depend not only on outdoor conditions and internal heat gains but also on zone layout, physical adjacency, and delayed thermal interactions across the building. Recent LLM-based HVAC controllers have shown that prompt-based control is feasible. However, these methods produce inconsistent results, are difficult to generalize, and experience control-latency issues. Their performance also relies heavily on numeric reward values, textual feedback, or unstructured retrieval, which limits their ability to make control decisions that account for zone dynamics and thermodynamics. No existing LLM-based methods have used building semantics and dynamics, together with rolling history, as the centre of the LLM context while testing and comparing HVAC control performance with existing LLM-based HVAC approaches. This paper presents ThermoLLM, an LLM-based HVAC control framework tested in a multi-zone building environment. Spatiotemporal context grounding is a main component of ThermoLLM, which uses a spatial knowledge graph derived from Brick-style building semantics and linked with recent observations from a rolling history. We evaluate ThermoLLM against standard control methods and recent state-of-the-art baselines in a five-zone building environment across six climate–season scenarios. ThermoLLM consistently achieves the best energy–comfort trade-offs and outperforms all evaluated baselines in the Technique for Order Preference by Similarity to Ideal Solution (TOPSIS) rankings across all six scenarios.\end{abstract}

\ccsdesc[500]{Information systems~Spatial-temporal systems}
\ccsdesc[500]{Computing methodologies~Ontology engineering}

\keywords{Large language models, multi-zone HVAC control, spatiotemporal context, building knowledge graphs, Brick schema, thermal comfort, energy efficiency.}

\usepackage{hyperref}
\usepackage{multirow}
\begin{document}\maketitle

{Code availability.} The Anonymised link: \url{https://github.com/inkwik/ThermoLLM.git}.

\section{Introduction}
Buildings account for a large share of global energy use, and HVAC systems remain one of the main drivers of that demand. At the same time, HVAC control must satisfy a difficult operational balance. It must reduce energy use while maintaining acceptable indoor comfort under changing weather, internal gains, occupancy patterns, and building-specific thermal behavior \cite{drgona2020all,serale2018model,maddalena2020datadriven}. The problem becomes harder in multi-zone buildings, where thermal coupling, heterogeneous zone responses, and building dynamics make fixed strategies and physics-agnostic controllers increasingly brittle. \cite{serale2018model,ding2024multizone}. Thermal conditions emerge from the arrangement of rooms, the placement of sensors and equipment, the connectivity of control points, and the physical coupling among adjacent zones. In multi-zone HVAC control, these spatial relationships matter directly. A control decision that is appropriate for one zone may be suboptimal for another because of different occupancy, solar exposure, equipment linkage, or thermal inertia. Yet most operational controllers still reduce the problem to local feedback or building-level optimization variables, with limited explicit use of building semantics and spatial structure.

Rule-based control is simple and widely deployed, but fixed heuristics adapt poorly to changes in weather, occupancy, and building-specific dynamics \cite{maddalena2020datadriven,wang2020reinforcement}. Model predictive control can optimize comfort and energy jointly, but it depends on model quality, cost function, and repeated online optimization \cite{serale2018model,drgona2020all}. Reinforcement learning can learn directly from interaction, but practical use is still constrained by sample cost, reward sensitivity, and weak transfer across environments \cite{ding2024multizone,manjavacas2024experimental}. These limits are especially visible in multi-zone settings, where heterogeneity and building dynamics raise both modeling and control complexity \cite{serale2018model,ding2024multizone}.

Large language models offer a different control substrate. Recent studies show that LLMs can directly choose HVAC actions from textualized observations without environment-specific training and can sometimes approach stronger learned baselines with much lower engineering overhead \cite{ahn2023alternative,song2023pretrained}. Recent task-specific interaction history can be more useful than generic demonstrations when control depends on local dynamics, as in HVAC \cite{song2023pretrained}.More recent building work also suggests that retrieval-augmented LLM control can improve interpretability and competitiveness by grounding decisions in external domain knowledge and historical operating patterns \cite{ko2025darlin}. These results are promising, but they also expose a clear gap. Existing LLM HVAC controllers are not consitent in their resulst and diffuslt to genralsie and heavily depend on numeric reward values, text feedback, or general control strategies that are not building-specific. Existing methods do not fully exploit the spatial organization of buildings as spatial-semantic context for LLM-based control. This context is important in multi-zone HVAC settings, where zone relationships, building structure, and thermal interactions can substantially influence control performance. Our comprehensive evaluation shows that incorporating such spatial-semantic awareness leads to improved multi-zone control outcomes.

A building is not only a thermodynamic object but also a spatial information structure: zones, sensors, setpoints, equipment, and control relationships can all be represented as entities and edges in a semantic graph. Recent LLM work outside HVAC shows that graph-guided and ontology-grounded context can improve reasoning quality by exposing multi-hop relations, constraining retrieval, and making context more faithful to domain structure \cite{ma2024think,sharma2024ograg,tian2024augmenting}. In physical-world sensor tasks, performance also depends strongly on how observations are structured, labeled, and filtered before being presented to the model \cite{an2025iotllm,ouyang2024llmsense}. Taken together, this literature suggests that the key issue is not only whether an LLM can reason about control actions but also whether it can better decide the control strategies for dynamic condition with right kind of relational and temporal context.

This paper addresses that issue through a spatiotemporal-aware control framework grounded in a spatio-temporal graph context. The setting is the standard five-zone EnergyPlus environment, where the controller must balance HVAC energy use and indoor comfort. The proposed method encodes building structure using Brick-style semantics so that the controller receives explicit information about zones, sensors, and their relations rather than only a flat observation vector and feedback. An important design choice is that these relations do not need to be manually written out as building-specific rules for every zone and equipment connection. Instead, the building configuration is automatically converted into a Brick-style graph through a Python conversion pipeline, allowing the controller to inherit a standardized and portable building representation from the simulation model itself \cite{balaji2016brick,teymourzadeh2025scalable}. In parallel, the controller receives a rolling window of recent environment interaction history, allowing it to observe short-term thermal response, comfort drift, and action consequences over time. The goal is to let the model reason over both where things are related in the building and how the building has recently responded.

The paper makes the following contributions:

\begin{itemize}

  \item We formulate multi-zone HVAC control as a spatial-semantic reasoning problem, where effective control requires awareness of building structure, zone coupling, and thermodynamic response.

  \item We propose a spatiotemporal context aware LLM control framework that combines a Brick-derived spatial knowledge graph with recent interaction history, allowing the controller to use both building structure and recent thermal trends when selecting zone-level setpoints.

  \item We demonstrate, through comparative experiments in a five-zone environment, that this context design improves the control performance relative to conventional controllers and LLM baselines.

\end{itemize}

Building control is a complex decision-making problem in which thermodynamic behavior plays a central role in selecting goal-aligned HVAC control strategies. The core claim of this paper is that the proposed framework enhances the thermodynamic reasoning capability of an LLM by grounding its decisions in spatial-semantic building context linked with recent interaction history. This enables the controller to reason not only from current observations, but also from building structure, zone-level relationships, and recent thermal response. The rest of the paper develops this argument through related work, method design, and comparative evaluation in the five-zone environment.

\section{Related Work}

\subsection{Baseline Control Methods for HVAC Control}

Rule-based control remains the most common practical baseline in building operation because it is simple, interpretable, and easy to deploy through existing building automation systems. Its main weakness is that fixed rules adapt poorly to changing weather, occupancy, and building-specific thermal behavior \cite{maddalena2020datadriven,wang2020reinforcement}. At the other end of the spectrum, model predictive control has become the strongest classical advanced baseline because it can explicitly optimize comfort and energy over a prediction horizon while enforcing operational constraints \cite{serale2018model,drgona2020all}. Reviews of MPC for buildings consistently show strong potential for energy savings and better use of thermal mass, but they also emphasize recurring deployment burdens: model identification, calibration effort, disturbance forecasting, and repeated online optimization. The mpc algorithm's performce heavily depend on model's prediction accuracy \cite{ding2020mb2c}

Reinforcement learning offers a different route by learning control policies directly from interaction rather than relying on an explicit control-oriented thermal model. Reviews and benchmark studies report strong promise for RL in HVAC control, especially in high-dimensional and multi-zone settings, but they also document persistent issues with sample efficiency, reward design, robustness, and transfer across buildings or operating conditions \cite{wang2020reinforcement,manjavacas2024experimental}. Classical Q-learning is especially difficult to scale to multi-zone HVAC control because the state and action spaces become large and continuous, forcing discretization that can reduce control precision and make learning sensitive to representation choices and operating conditions \cite{wang2023comparison}. More recent multi-zone work shows that model-based deep RL can improve data efficiency and energy savings in five-zone EnergyPlus environments, which makes RL an important comparison point for modern building-control studies \cite{ding2024multizone}. recent work used diffusion model with RL show early and better convergence of reward and better generalisation with addition of physics constariend model training \cite{yang2026adapt}. how Taken together, rule-based control, MPC, and RL represent the main baseline families against which new HVAC controllers are usually judged: practical heuristics, optimization-based control, and learned adaptive policies \cite{serale2018model,maddalena2020datadriven,wang2020reinforcement}.

\subsection{LLM-based Control Works}

The building-control literature has also begun to shift from zero-shot prompting toward context-grounded LLM control. \cite{ko2025darlin} proposed a retrieval-augmented HVAC controller that combines real-time operational states with retrieved domain literature and historical operating patterns, demonstrating that external grounding can improve both controller competitiveness and interpretability. Similarly, \cite{sawada2025office} demonstrated a real-world implementation of LLM control using a multimodal foundation model in an office building, reporting that the LLM-based approach reduced energy consumption by 47.9\% and decreased comfort dissatisfaction by 26.3\% compared to the baseline. Notably, their study indicated that incorporating building thermal imagery did not significantly improve the energy-comfort trade-off. This suggests that appending redundant context can fail to enhance performance, and may instead degrade decision-making reasoning, leading to suboptimal control policies. Along the same lines, \cite{song2023pretrained} showed that introducing expert demonstrations as context alongside operational history led to diminishing performance, whereas utilizing history alone yielded superior results. Together, these studies establish that LLM-based HVAC control is already highly feasible. However, current methods still rely primarily on textual observations, numeric rewards, text feedback,  demonstrations, or unstructured retrieval, also they did not show the llm's consistency in results with multiple runs, tetsing diverese scenarios and they did not accont the tetsing with diffemre t models,latency and cost during the operations. They do not yet position spatial building semantics at the center of the control context, nor do they explicitly treat zone relationships, equipment topology, and building ontologies as first-class inputs for controller reasoning.

\subsection{Knowledge Graphs and Building Ontologies}

Building ontologies provides a natural basis for representing HVAC control as a spatial-semantic problem. Brick was introduced to address the fragmentation of vendor-specific building metadata by defining a common ontology over locations, equipment, points, and their relations, enabling portable applications and graph-based querying across buildings \cite{balaji2016brick}. This matters for HVAC control because Brick does not only name components; it also captures the structural relations among rooms, floors, sensors, setpoints, and HVAC equipment that determine how control decisions propagate through a building. More recent work has shown how heterogeneous building information and point lists can be transformed into Brick-based graph representations at scale, improving interoperability and making graph-structured building context more usable in digital twin workflows \cite{teymourzadeh2025scalable}. other relevant brick conversion approaches  and studies are \cite{granderson2023labeled, zheng2023brick}.

Parallel work on LLM grounding suggests that graph and ontology structure can improve how models use external context. In building applications, retrieval-augmented LLMs have already been used to ground automated building-energy modeling in historical structured references rather than relying on parametric knowledge alone \cite{zhang2026large}. More generally, graph-guided and ontology-grounded retrieval methods show that LLM reasoning and performance improves when context is organized around typed entities and relations rather than only flat text similarity \cite{ma2024think,sharma2024ograg,tian2024augmenting}. \cite{ma2024think} shows that knowledge graphs can act as a navigation structure for deeper retrieval and more consistent reasoning, while \cite{tian2024augmenting} demonstrates the value of multi-hop relational paths as prompt context. \cite{sharma2024ograg} is especially relevant here because it argues that ontology-grounded retrieval is better suited to domains where correct relations and procedural structure matter. 

These findings motivate the use of a Brick-derived spatial knowledge graph in the proposed controller. The graph explicitly represents relevant spatial, equipment, and sensing relationships, providing structured building context alongside current observations and rolling interaction history. Since LLM outputs are sensitive to prompt templates and wording \cite{chatterjee2024posix}, a standardized context representation is an important design consideration. Mapping different buildings to a common Brick schema allows the framework to reuse the same retrieval logic and prompt template while populating them with building-specific entities. This produces a uniform spatiotemporal context structure, reduces reliance on manually maintained descriptions, unstructured retrieval, and simplifies adaptation across buildings.

\section{Methodology}

\subsection{Framework Overview}

\begin{figure}[t]
    \centering
    \begin{tikzpicture}[baseline]
    \node[draw=black, dotted, line width=0.55pt, inner sep=3pt, outer sep=0pt] {\imwutimage{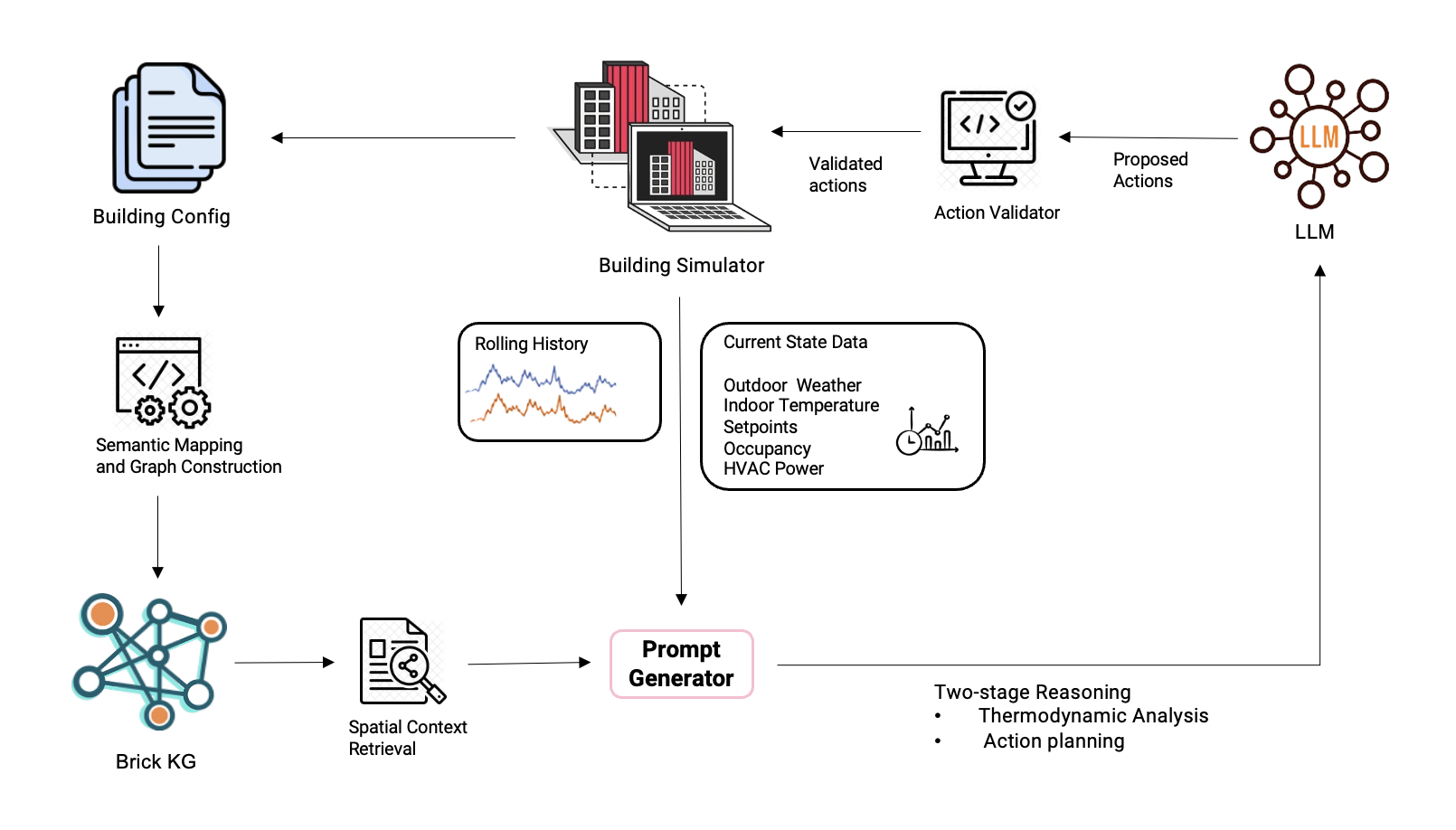}{0.90}};
\end{tikzpicture}
    \caption{Overall Framework of ThermoLLM: Spatial-Semantic Knowledge Graph Construction and LLM-Based HVAC Control Loop}
    \Description{Framework schematic of the spatial-semantic knowledge graph, building observations, recent control history, large language model, and HVAC control loop.}
    \label{fig:framework}
\end{figure}

The proposed framework treats multi-zone HVAC control as a context-grounded reasoning loop in which the controller receives three information sources at each simulation step: the current building state, spatial-semantic context retrieved from a building knowledge graph, and a rolling history of recent environment-controller interactions. The overall workflow is shown in the Figure~\ref{fig:framework}  First, the five-zone building configuration is converted into a Brick-style knowledge graph. Second, during simulation, the current observation is collected from the EnergyPlus environment. Third, zone-relevant graph context and recent interaction history are assembled into a structured prompt for the LLM. The LLM then generates heating and cooling setpoints for all five zones; the action is validated against control constraints, and the simulator executes the action. The resulting thermal and energy response is appended to the rolling history, which becomes part of the context at the next control step.

This design is intended to make the controller both spatially and thermodynamically aware. The spatial-semantic layer provides explicit information about how each zone is situated within the building control structure, including its linked thermostat, sensors, VAV reheat unit, and other zone-level entities represented in the building graph. The recent-history layer provides short-horizon evidence about how temperatures, weather, and power use have evolved under recent actions. The controller therefore reasons not only from the present observation but also from building structure and recent thermal response.

\subsection{Multi-Zone Building Simulation and Control Problem}

Experiments are conducted in the five-zone office building environment provided through Sinergym and simulated with EnergyPlus \cite{campoynieves2024sinergym}. The building is a single-floor office model with five thermal zones: four perimeter zones and one core zone. Each zone is controlled independently through its heating and cooling setpoints, which makes the task a continuous multi-zone supervisory control problem rather than a single global thermostat adjustment. EnergyPlus is used because it is a widely adopted building-energy simulation engine and provides physically detailed zone-level thermal dynamics, HVAC behavior, and weather-driven responses. Within Sinergym, the environment exposes observations and accepts actions through a control loop suitable for reproducible controller comparison.

The controller acts at a 15-minute timestep. At each step, the observation includes zone-wise indoor air temperature, indoor humidity, occupancy for each zone, the previously applied action, outdoor air temperature, direct solar radiation, aggregate HVAC power consumption, and time indicators. The action space consists of a heating setpoint and a cooling setpoint for each of the five zones, giving ten continuous control values per step. The allowable setpoint bounds are $12.0\,^{\circ}\mathrm{C}$ to $23.25\,^{\circ}\mathrm{C}$ for heating and
$23.25\,^{\circ}\mathrm{C}$ to $30.0\,^{\circ}\mathrm{C}$ for cooling. The control objective is to select zone-level setpoints that reduce occupied-zone PMV-based discomfort and HVAC energy use while respecting the actuator limits of the environment.

\subsection{Spatial-Semantic Knowledge Graph Construction and Retrieval}

To provide the controller with explicit building structure, the five-zone office building is converted into a Brick-style knowledge graph \cite{balaji2016brick} using a custom Python conversion pipeline. We use an approach to Brick conversion similar to those used in prior work \cite{granderson2023labeled, zheng2023brick}. First, the pipeline parses the building configuration file to extract individual zones, sites, and building points, defining them as distinct graph entities and establishing the foundational relationships and lookup connections between these components. Next, these entities and their structural relations are mapped directly to standard Brick vocabulary and semantic labels using the schema's semantic mapping rules, culminating in the final generated Brick ontology graph. This point is important for the scope and portability of the method. The framework does not require manually writing detailed building-specific rules for every zone connection, sensor relation, or HVAC path. Instead, the simulation configuration is automatically mapped into standardized graph entities and relations, so that the building structure becomes available to the controller through a common semantic schema rather than through hand-written descriptions. Brick is well suited to this role because it was designed to standardize the semantic description of physical structural assets in buildings together with their relationships. This makes the graph easier to query, easier for an LLM to interpret as structured context, and more portable to other buildings whose models can be converted into the same schema. A recent study introduces a scalable process to convert any building configuration into a Brick ontology using only historical data and a building point list \cite{teymourzadeh2025scalable}. This approach is also applicable to real-world buildings that lack existing floor plans or models.

At runtime, the full building graph is not passed to the LLM directly. Instead, a retrieval step extracts the most relevant structural context using the RDFLib library. Retrieval is performed per zone and returns a compact subgraph focused on that zone's important connections and shared HVAC relations. In practice, this includes the zone type and label, its relations, incoming links from sensors and setpoints, thermostat-to-zone control relations, VAV-to-zone feed relations, and one-hop neighborhood triples that summarize the local structural context. A typical retrieved block therefore states, for example, that a zone contains a thermostat and a VAV reheat unit, is associated with heating and cooling setpoint objects, receives air from a specific VAV, and is linked to a zone temperature sensor. In addition to zone-specific subgraphs, common HVAC-unit information is retrieved so that the prompt preserves the shared supply-side structure of the building. This retrieval design keeps the context compact while preserving the physically meaningful relations most likely to affect control.

\subsection{Spatiotemporal Context and Prompt Generation}

The second contextual component is a rolling interaction history over the previous 10 simulation steps. The purpose of this history is not to estimate a formal thermal model, but to expose short-term response patterns that a single observation cannot reveal. For each step in the history window, the stored record includes the zone temperatures, zone occupancies, solar radiation, previous heating and cooling setpoints, outdoor conditions, aggregate HVAC power consumption, and time variables. This gives the controller a short temporal trace of how the building has recently evolved under prior actions. Recent literature on LLM-based HVAC control demonstrates that integrating a short historical data block provides meaningful thermodynamic context, enabling the model to effectively reason over temporal trends and optimize subsequent control actions\cite{ko2025darlin}. Brick supports external time-series references that associate sensor points in the knowledge graph with their corresponding data streams stored in an external database. We use this mechanism to link sensor data to the building entities and relationships represented in the graph. Combining the retrieved measurements with their spatial and semantic context provides structured spatiotemporal context for the LLM, supporting reasoning about zone-level thermal behavior and its evolution under changing conditions.

The prompt at each control step combines three blocks: the current observation, the retrieved spatial-semantic graph context, and the history window. The current observation tells the model what the building state is now. The graph context tells the model how each zone is embedded in the building's control structure. The history tells the model how indoor and outdoor temperature trends and power use have been moving over time. Together, these blocks are meant to help the LLM reason about thermal inertia, persistent comfort drift, power jumps, and differences among zone behaviors and outdoor influeance and action effect. For example, if a perimeter zone with large window has been warming steadily under increasing solar load despite previous cooling actions, while the graph context links that zone to a specific thermostat and VAV path, the model receives the structural, relational, and temporal evidence needed to adjust its action more deliberately.

\subsection{LLM-Based Control Action Generation}

At each 15-minute control interval, the LLM receives the assembled prompt and generates a structured heating and cooling setpoint plan for each of the five zones over a 75-minute horizon. The planning task encourages the model to consider delayed thermal responses and the anticipated comfort and energy implications of its decisions. The controller follows a receding-horizon execution strategy: only the setpoints for the immediate 15-minute interval are applied. At the next observation, the context is updated and a new plan is generated over the subsequent 75-minute horizon. The complete response is parsed and validated before execution. 

Every proposed setpoint pair is checked to ensure that $12.0\,^{\circ}\mathrm{C} \leq HSP_i \leq 23.25\,^{\circ}\mathrm{C}$, $23.25\,^{\circ}\mathrm{C} \leq CSP_i \leq 30.0\,^{\circ}\mathrm{C}$, and $HSP_i < CSP_i$. Responses with malformed JSON, missing or invalid values, or constraint violations are rejected. Following validation, the immediate setpoint pair for each zone is extracted to form the ten-dimensional control action. An invalid response immediately activates rule-based fallback, without an additional LLM request to correct or regenerate the plan. The same fallback is used if the API request fails. Fallback setpoints are clipped to the permitted bounds and corrected to maintain strict heating–cooling ordering, ensuring that only constraint-compliant actions are submitted to the simulator. Throughout execution, we record API latency, token usage, estimated API cost, request and validation failures, and the action source at each interval, enabling calculation of fallback and LLM-action application rates.

Once validated, the action is applied to the EnergyPlus environment through the Sinergym control interface. The simulator then advances one 15-minute step and returns the next state, including updated zone conditions and HVAC power use. These responses are recorded and appended to the rolling history buffer, which drops the oldest entry once the history length exceeds 10 steps. The same process then repeats. This rolling control loop allows the controller to continually refresh its thermodynamic context while keeping the prompt grounded in a compact, zone-relevant slice of the building graph and the most recent trajectory information.

\section{Experimental Setup and Implementation}

\subsection{Simulation Environment}

The proposed controller is evaluated in the five-zone office building environment implemented through Sinergym and simulated with EnergyPlus \cite{campoynieves2024sinergym}. The building consists of four perimeter zones and one core zone on a single floor, and each zone has its own heating and cooling setpoint action. The environment is therefore a multi-zone supervisory control problem in which the controller must coordinate ten continuous control values at each decision step. Consistent with the methodology described above, the controller is invoked every 15 minutes, and the resulting action is executed for one simulation step before the next observation is collected. We have used real occupancy data \cite{tekler2022robod} instead of fixed time based occupancy schedule given in the simulation to have realistic occupancy pattern context and control. We constructed zone-specific occupancy schedules from ROBOD’s five-minute ground-truth occupant counts using a fixed one-to-one assignment between ROBOD's source rooms and simulated zones

To evaluate ThermoLLM under realistic and diverse outdoor conditions, we consider six climate–season scenarios spanning Arizona, New York, and Chicago, with summer and winter periods at each location. Simulations use location-specific Typical Meteorological Year (TMY3) weather datasets through Sinergym. Constructed from historical weather records, these datasets capture representative variations in temperature, humidity, solar radiation, and wind. This experimental design assesses the controller’s robustness and adaptability across varied climatic conditions. 

\begin{figure}[tbp]
    \centering
    \imwutimage{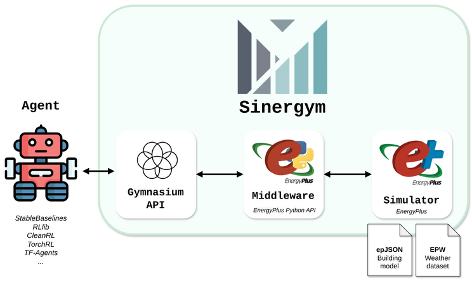}{0.80}
    \caption{Sinergym: EnergyPlus-Based Simulation Environment}
    \Description{Illustration of the Sinergym and EnergyPlus Simulation Environment used to evaluate five-zone HVAC control.}
    \label{fig:sinergym}
\end{figure}

\subsection{Compared Controllers}

The proposed framework is evaluated against following baseline control methods:

\subsubsection{Rule-Based Control:} The rule-based controller (RBC)for the 5-zone office building follows a seasonal setpoint schedule derived from ASHRAE 55-2004 (cooling 23°C/heating 26°C in summer; cooling 20°C/heating 23.5°C in winter), as implemented in the Sinergym framework \cite{sinergym2025}.

\subsubsection{Reinforcemnrt Learning Based Control:}
We evaluate model-free reinforcement learning baselines: Proximal Policy Optimization (PPO) \cite{ding2024multizone} trained with a reward function, as shown in equation ~\ref{eq:reward_fnc}. PPO is trained with the same reward function, defined as the negative weighted sum of normalized HVAC energy consumption and occupancy-weighted comfort violations across the five zones. The energy and comfort terms are normalized based on empirical ranges observed in the training logs to improve reward scaling and stabilize learning. The occupancy-dependent comfort weight is denoted by $\rho_i$: for occupied periods, we set $\rho_i=1.1$ to prioritize comfort preservation, while for unoccupied periods, we set $\rho_i=0.1$ to allow energy-saving setbacks. For robust and strong ppo basleine we trained ppo model seperatly for each location and season with stochastic enviroennt and tested on 1 complete week of winter and summer periods for each locations. following are the ppo setitng used, two 200-unit hidden layers and ReLU activation, learning rate 0.0003; batch size 512; 5 epochs per update; $\gamma$=0.99; GAE $\lambda$=0.95; clip range 0.1; target KL 0.015; entropy coefficient 0.001; value coefficient 0.5; max gradient norm 0.5.

The reward function is defined as follows:
\begin{equation}
\label{eq:reward_fnc}
R = -\left( \sum_{i=1}^{5} \rho_{i} \cdot \mathrm{Norm}(PMV_{i}) + \mathrm{Norm}(P_{\mathrm{HVAC}}) \right),
\end{equation}

\noindent where $\rho_{i}$ represents the occupancy-weighted comfort scaling factor.

\subsubsection{Mode Based Control:}
MB2C learns an ensemble of neural networks to predict how the building responds to control actions, then uses model predictive control (MPC) to plan HVAC setpoints over a future horizon. At each step, MPPI samples candidate heating and cooling setpoint sequences, scores their predicted comfort and energy costs, and applies the best action before replanning from the next observation.- Predictive model: five-member neural-network ensemble, trained on six years of data across Arizona, Chicago, and New York.
- Planning: MPPI with a 20-step horizon (5 hours), 1,000 candidate action sequences, and 5 optimization iterations per control step. \cite{ding2024multizone}. we tetsed this on same duration and setup as other balines methods.

\subsubsection{LLM-Based Control:}
DARLIN-Domain-guided Augmented Retrieval for LLM-based INterpretable HVAC Control, is a training-free, zero-shot HVAC control framework that combines LLM with Retrieval-Augmented Generation (RAG) It optimizes multi-zone cooling setpoints without requiring building-specific system modeling or reinforcement learning training. To replicate this for our 5zone building expermints we used the same 5,092 domain-specific research papers data base and other setup provided in paper and repository. \cite{ko2025darlin, yun2026darlin}. We used a 15-minute control interval instead of DARLIN's original 30-minute interval to ensure a fair comparison with ThermoLLM.

\subsection{Implementation Details of ThermoLLM}

As explained in Section 4.2, the knowledge graph construction and retrieval process has already been described. Here, we focus on what happens at each simulation step and how the LLM determines the control action. We used GPT Sol-6 through the OpenAI API across all runs, with the LLM temperature set to 0.0 and reasoning effort set to none. We also used the same role, goal, output schema, and LLM model settings for DARLIN to ensure a fair comparison with our framework. The prompt structure is fixed across experiments. The system prompt defines the controller role, control objective, action bounds, safety constraints, causality notes, and the required output format. The user prompt then provides the current observation, the rolling history block, and the retrieved knowledge graph context. The history grows over time to a maximum of 10 steps and is maintained on a rolling basis, retaining only recent interaction history in the context.

Once the LLM receives the context, it performs two-stage reasoning: first, a thermodynamic analysis of the spatiotemporal context; and second, future-impact-aware action planning, proposing a plan for the next five steps. The output schema is strictly defined to require a five-step plan and a thermodynamic and goal-oriented justification for each zone. Only the first action in the proposed plan is applied. At the next control step, the LLM receives updated observations and generates a new five-step plan. This process continues throughout the evaluation period. The value of this graph-linked context is further reflected in the results section.

A safety-validation layer checks every generated action before simulator execution. If any heating or cooling setpoint falls outside the allowable bounds or an API call fails, the controller applies the predefined safe fallback rules. The controller is informed of both the generated action and the action actually applied so that it can adjust its subsequent decisions when a fallback occurs. Fallback activations, the LLM action application rate, latency, and cost are recorded throughout operation. After each simulation step, the resulting comfort response, energy consumption, selected action, and zone-level observations are recorded and appended to the rolling history used at the next step. This control loop continues over the full evaluation period. Thermal comfort statistics and energy consumption are then computed to support a comprehensive evaluation and comparison of all methods.

\section{Results}

We evaluate all controllers in the same five-zone environment over a one-week period across six climate--season scenarios. Performance is assessed using two main metrics: total HVAC energy consumption in kilowatt-hours (kWh) and the predicted mean vote (PMV) violation rate. Energy use is computed as the cumulative HVAC consumption over the full evaluation period. The PMV violation rate measures the proportion of occupied samples that fall outside the acceptable thermal comfort range. PMV is calculated based on \cite{fanger1970thermal}. The six variables used in the PMV calculation are air temperature ($T_a$), mean radiant temperature ($T_r$), air velocity ($v$), relative humidity ($RH$), clothing insulation ($I_{cl}$), and metabolic rate ($M$). We assume a relative humidity of $RH = 40\%$, an air velocity of $v = 0.15\,\mathrm{m/s}$, and a metabolic rate of $M = 1.1\,\mathrm{met}$. Mean radiant temperature ($T_r$) is approximated as equal to the indoor zone air temperature ($T_a$), and clothing insulation is set to $0.5\,\mathrm{clo}$ for summer and $1.0\,\mathrm{clo}$ for winter. These assumptions are based on \cite{ashrae2023standard55}. This ensures that thermal comfort improvements directly reflect the HVAC controller's setpoint and temperature management. The acceptable PMV range is $[-0.5, +0.5]$. Together, these metrics capture the central trade-off in HVAC control: reducing discomfort without incurring unnecessary energy costs.

Since energy consumption and thermal comfort are competing objectives, we analyze controller performance using a Pareto comfort--energy trade-off plot \cite{ascione2016simulation} and the TOPSIS metric \cite{mao2016application}.

\subsection{Pareto plot Analysis}

\begin{figure*}[t]
  \centering
  \includegraphics[width=\textwidth]{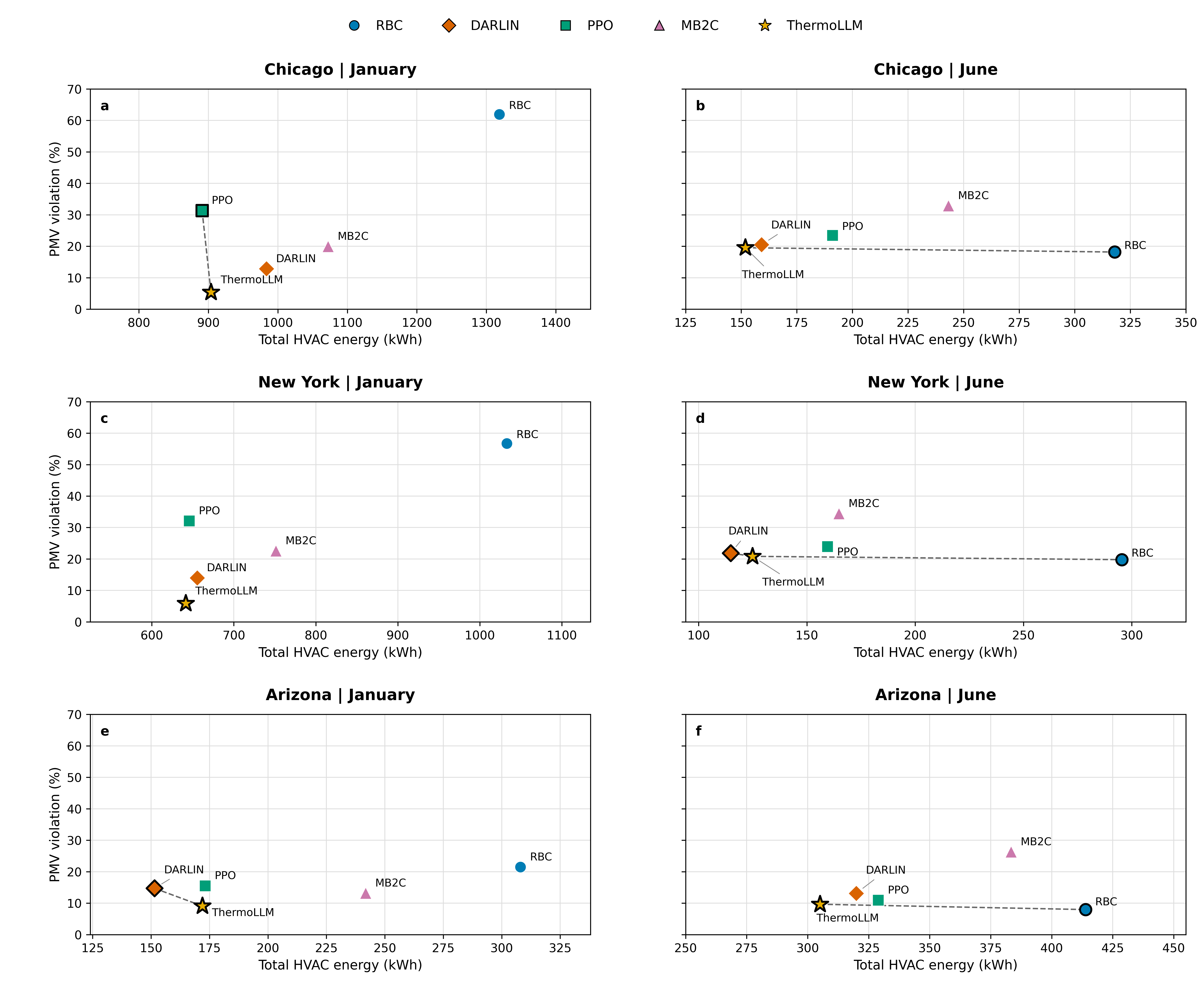}
  \caption{Pareto scatter comparison across Chicago, New York, and Arizona for January and June.}
  \label{fig:pareto_six_scenarios}
\end{figure*}

A Pareto plot shows the trade-off between competing objectives, here HVAC energy use and comfort violations. The lower-left region is preferable because both values are lower. A method is on the Pareto front if no other evaluated method is at least as good on both metrics and strictly better on one. Figure~\ref{fig:pareto_six_scenarios} compares the five controllers across six climate--season scenarios using the aggregate PMV violation rate and total HVAC energy consumption. Both objectives are minimized, so points closer to the lower-left corner indicate a more favorable energy--comfort trade-off. The plot shows that ThermoLLM consistently achieves the best energy--comfort trade-off, lying on the Pareto front across all six climate--season scenarios. It achieves lower HVAC energy consumption and fewer PMV violations than MB2C in every scenario, and than PPO and DARLIN in five and four scenarios, respectively.

\subsection{TOPSIS Ranking}

\begin{table}[H]
  \centering
  \caption{Scenario-wise TOPSIS rankings using aggregate energy and PMV violation. Each entry reports rank ($C_i$); rank 1 is best.}
  \label{tab:topsis_results}
  \small
  \setlength{\tabcolsep}{3.5pt}
  \renewcommand{\arraystretch}{1.15}
  \begin{tabular}{@{}lccccccc@{}}
    \hline
    \textbf{Controller} & \shortstack{Chicago\\Jan.} & \shortstack{Chicago\\Jun.} & \shortstack{New York\\Jan.} & \shortstack{New York\\Jun.} & \shortstack{Arizona\\Jan.} & \shortstack{Arizona\\Jun.} & \shortstack{Avg.\\rank} \\
    \hline
    RBC          & 5 (0.000) & 4 (0.455) & 5 (0.000) & 5 (0.376) & 5 (0.000) & 3 (0.798) & 4.50 \\
    MB2C         & 3 (0.732) & 5 (0.311) & 3 (0.675) & 4 (0.522) & 4 (0.557) & 5 (0.066) & 4.00 \\
    PPO          & 4 (0.562) & 3 (0.709) & 4 (0.529) & 3 (0.742) & 3 (0.629) & 2 (0.834) & 3.17 \\
    DARLIN       & 2 (0.862) & 2 (0.895) & 2 (0.848) & 2 (0.929) & 2 (0.697) & 4 (0.727) & 2.33 \\
        ThermoLLM    & \textbf{1 (0.993)} & \textbf{1 (0.942)} & \textbf{1 (1.000)} & \textbf{1 (0.940)} & \textbf{1 (0.916)} & \textbf{1 (0.911)} & \textbf{1.00} \\
    \hline
  \end{tabular}
\end{table}

To complement the Pareto analysis with a combined assessment of energy and comfort, we apply the Technique for Order Preference by Similarity to Ideal Solution (TOPSIS) separately to each scenario. Total HVAC energy consumption and the PMV violation rate are vector-normalized across the evaluated controllers and each assigned a weight of 0.5. We then calculate each controller's Euclidean distances to the ideal best and worst combinations, defined by the minimum and maximum weighted normalized values, respectively. Controllers are ranked in descending order of the closeness score, $C = D_{\mathrm{worst}}/(D_{\mathrm{best}} + D_{\mathrm{worst}})$.

In Table~\ref{tab:topsis_results}, values outside parentheses indicate scenario-specific ranks, with rank 1 representing the best combined energy--comfort performance under the chosen weighting. Values in parentheses are TOPSIS closeness scores ranging from 0 to 1, with higher scores indicating greater relative proximity to the ideal combination of low energy consumption and low PMV violation. The average rank summarizes performance across all six scenarios, with lower values indicating better overall ranking. ThermoLLM ranks first in all six climate--season scenarios under this equal-weight assessment. DARLIN ranks second in five scenarios, while PPO ranks second in Arizona in June. The remaining baselines exhibit greater variation in rank across scenarios. These results indicate that ThermoLLM consistently provides the most favorable combined energy--comfort performance.

\subsection{Thermal Comfort Statacstics}

\begin{table}[H]
  \centering
  \caption{Thermal comfort statistics across controllers and climates.}
  \label{tab:thermal_comfort_statistics}
  \small
  \setlength{\tabcolsep}{3.2pt}
  \renewcommand{\arraystretch}{1.18}
  \resizebox{\textwidth}{!}{%
  \begin{tabular}{|c|c|c|*{10}{c|}}
    \hline
    \multirow{2}{*}{\textbf{Location}} & \multirow{2}{*}{\textbf{Comfort}} & \multirow{2}{*}{\textbf{Metric}}
      & \multicolumn{2}{c|}{\textbf{RBC}}
      & \multicolumn{2}{c|}{\textbf{MB2C}}
      & \multicolumn{2}{c|}{\textbf{PPO}}
      & \multicolumn{2}{c|}{\textbf{DARLIN}}
      & \multicolumn{2}{c|}{\textbf{ThermoLLM}} \\
    \cline{4-13}
    & & & \textbf{January} & \textbf{June}
      & \textbf{January} & \textbf{June}
      & \textbf{January} & \textbf{June}
      & \textbf{January} & \textbf{June}
      & \textbf{January} & \textbf{June} \\
    \hline
    \multirow{3}{*}{Chicago} & \multirow{3}{*}{PMV} & Mean      & -0.479 & -0.203 & -0.251 & -0.190 & -0.253 & -0.032 & -0.296 & -0.080 & -0.259 & -0.005 \\
    \cline{3-13}
    & & Std      & 0.362 & 0.348 & 0.366 & 0.486 & 0.528 & 0.445 & 0.439 & 0.453 & 0.365 & 0.460 \\
    \cline{3-13}
    & & Violation (\%)      & 61.95 & 18.18 & 19.95 & 32.90 & 31.38 & 23.49 & 12.87 & 20.51 & 5.39 & 19.55 \\
    \hline
    \multirow{3}{*}{New York} & \multirow{3}{*}{PMV} & Mean      & -0.425 & -0.218 & -0.266 & -0.227 & -0.229 & -0.072 & -0.239 & 0.091 & -0.201 & -0.025 \\
    \cline{3-13}
    & & Std      & 0.391 & 0.361 & 0.464 & 0.490 & 0.488 & 0.457 & 0.463 & 0.532 & 0.381 & 0.469 \\
    \cline{3-13}
    & & Violation (\%)      & 56.72 & 19.79 & 22.61 & 34.43 & 32.10 & 23.97 & 14.00 & 21.88 & 5.87 & 20.84 \\
    \hline
    \multirow{3}{*}{Arizona} & \multirow{3}{*}{PMV} & Mean      & -0.050 & -0.079 & 0.137 & -0.127 & -0.013 & 0.124 & 0.199 & 0.150 & 0.178 & 0.178 \\
    \cline{3-13}
    & & Std      & 0.374 & 0.248 & 0.358 & 0.418 & 0.364 & 0.314 & 0.475 & 0.381 & 0.384 & 0.346 \\
    \cline{3-13}
    & & Violation (\%)      & 21.48 & 7.96 & 13.19 & 26.31 & 15.53 & 10.94 & 14.72 & 13.11 & 9.09 & 9.65 \\
    \hline
  \end{tabular}%
  }
  \vspace{2pt}
  
\end{table}

Table~\ref{tab:thermal_comfort_statistics} summarizes thermal comfort using the mean, standard deviation, and violation rate of PMV across occupied zone--time samples. Following the PMV-based comfort criterion in ASHRAE Standard 55, values within $[-0.5, +0.5]$ are considered acceptable. Values below $-0.5$ indicate sensations on the cool side of this range, while values above $+0.5$ indicate sensations on the warm side. PMV values closer to zero indicate predicted thermal sensations closer to neutrality. Mean PMV describes the average thermal sensation, while the standard deviation describes variation across samples. The violation rate measures the proportion of samples outside the PMV comfort interval $[-0.5, +0.5]$. Reporting these metrics together provides a fuller assessment of comfort performance, capturing average thermal bias and variation across occupied samples while identifying which controllers achieve average thermal sensations closer to neutrality. The three mean PMV values closest to neutrality are achieved by ThermoLLM in Chicago in June ($-0.005$), PPO in Arizona in January ($-0.013$), and ThermoLLM in New York in June ($-0.025$). RBC in Chicago in January has a mean PMV of $-0.479$, approaching the lower comfort boundary of $-0.5$. The study in \cite{ding2024multizone} uses a PMV comfort interval of $[-0.7, +0.7]$, considering 85\% thermal satisfaction, which is why its PMV violation percentages are lower than those in our statistical evaluation of PMV.

\subsection{Operational Efficiency, Reliability, and Consistency of LLM based  Controller}

  \begin{table}[H]
  \centering
    \caption{Operational Cost, Latency, and Reliability.}
  \label{tab:operational_cost}
  \small
  \setlength{\tabcolsep}{8pt}
  \renewcommand{\arraystretch}{1.2}
  \begin{tabular}{@{}lcc@{}}
    \hline
    \textbf{Metric} & \textbf{DARLIN} & \textbf{ThermoLLM} \\
    \hline
        API cost (USD) & $21.584 \pm 1.541$ & $13.251 \pm 0.540$ \\
    API failure rate (\%) & $0.0000 \pm 0.0000$ & $0.0000 \pm 0.0000$ \\
    API-request latency (s) & $6.220 \pm 0.849$ & $4.084 \pm 0.305$ \\
    Fallback rate (\%) & $0.1181 \pm 0.0881$ & $0.0744 \pm 0.0245$ \\

    \hline
  \end{tabular}
\end{table}

Table~\ref{tab:operational_cost} compares operational cost, API-request latency, and reliability across seven runs covering six climate--season scenarios. These metrics assess the computational overhead and operational dependability of the LLM-based controllers alongside their energy--comfort performance. DARLIN uses separate LLM requests to summarize retrieved literature information and generate control actions, increasing its request overhead. ThermoLLM incurs a lower average API cost per seven-day run and lower API-request latency, corresponding to reductions of approximately 38.6\% and 34.3\%, respectively. Both methods record zero API failures. Fallbacks are activated when the LLM output fails format or setpoint validation. ThermoLLM records a lower fallback rate, indicating infrequent reliance on fallback control. Together with the energy--comfort results, these findings support ThermoLLM's operational efficiency and reliable execution under the evaluated conditions.

\begin{table}[H]
  \centering
  \caption{Energy--Comfort Consistency.}
  \label{tab:energy_comfort_consistency}
  \small
  \setlength{\tabcolsep}{8pt}
  \renewcommand{\arraystretch}{1.2}
  \begin{tabular}{@{}lcc@{}}
    \hline
    \textbf{Metric} & \textbf{DARLIN} & \textbf{ThermoLLM} \\
    \hline
    HVAC energy (kWh) & $335.04 \pm 35.61$ & \textbf{$304.80 \pm 0.27$} \\
    PMV violation rate (\%) & $11.76 \pm 2.10$ & \textbf{$9.11 \pm 0.36$} \\
    \hline
  \end{tabular}
\end{table}

Table~\ref{tab:energy_comfort_consistency} evaluates the repeatability of energy--comfort performance across five seven-day Arizona June runs using the same LLM model and inference settings for both controllers. Reporting the mean and standard deviation across repetitions establishes whether performance is consistently reproduced rather than relying on a single evaluation. ThermoLLM achieves lower mean HVAC energy consumption than DARLIN and a lower PMV violation rate across all five runs. These correspond to a 9.03\% reduction in energy consumption and a 2.65-percentage-point reduction in comfort violations. ThermoLLM also exhibits substantially smaller standard deviations for energy (0.27 versus 35.61 kWh) and violation rate (0.36 versus 2.10 percentage points). These results demonstrate better average performance and stronger repeatability in this scenario, supporting more predictable and practical control outcomes across repeated executions.

\section{Discussion}

The results suggest that the main advantage of ThermoLLM is not simply the use of graph context, but the use of decision-relevant spatial-semantic context at the zone level linked with rolling history. This context helps the LLM make more informed control decisions in a multi-zone setting. Another major benefit evident from the results is that ThermoLLM consistently performs well across all six scenarios without requiring training or fine-tuning. In contrast, PPO requires years of training to adapt to new conditions. MB2C is data-efficient, but requires training an accurate model using high-quality data. Defining an appropriate reward function, selecting optimal comfort and energy weights, and tuning hyperparameters are additional ad hoc and challenging tasks for both methods. Rule-based control is easy to define for new conditions, but, as the results show, it performs worst across all cases because it cannot account for the five-zone dynamics and balance multi objective well. Defining stronger and more detailed rules for every condition across different scenarios is difficult.

The TOPSIS ranking table shows that DARLIN performs well compared with the non-LLM baselines and ranks second overall. DARLIN also achieves the lowest energy consumption in New York in June and Arizona in January compared with all other baselines and ThermoLLM, but at the cost of higher PMV violation rates than ThermoLLM. Tables 3 and 4 highlight DARLIN's limitations regarding operational overhead and consistency. Like ThermoLLM, DARLIN requires no training and can be readily adapted to different conditions. However, its performance depends heavily on the quality of the literature database and the literature summaries generated by the LLM at each control step. As discussed, LLM outputs are sensitive to input context \cite{chatterjee2024posix}. The quality and content of these literature summaries affect performance across repeated runs of the same scenario. As shown in Table 4, DARLIN exhibits substantially less consistent performance than ThermoLLM. ThermoLLM consistently produces similar results because its input context remains consistent and structured throughout operation and across multiple runs. Another consideration for LLM-based controllers such as ThermoLLM and DARLIN is that output token counts depend on the server during computation, even when reasoning is disabled and the LLM temperature is set to zero. However, this difference is not significant.

The results and comparisons with other baseline methods suggest that spatiotemporal context helps the model interpret recent thermal behavior in a more localized and structurally grounded way. Instead of reasoning only from recent trends, the controller can relate those trends to zone-level building context, including spatial structure, control relationships, and building semantics. One could argue that spatial context could instead be added directly as manually written prompt descriptions or notes. In the DARLIN approach, however, this spatial information is already included in the prompt instructions, following the authors' implementation. As the results show, the performance gap between DARLIN and ThermoLLM therefore supports the claim that the form of context matters, not only its amount. Furthermore, manually writing these descriptions and maintaining a consistent structure pose additional challenges, as discussed earlier. In particular, structured spatial-semantic context appears to be more useful than additional static prompt guidance, expert demonstrations, or generic control strategies retrieved from literature.

Evaluating six diverse scenarios was important for robust testing, as the results show that no single method achieves both the lowest violation rate and the lowest energy consumption in every scenario. Changes in climate and outdoor conditions affect performance and test controller adaptability. Among the conventional baselines, RBC maintains the lowest PMV violation rate in all June scenarios, but at the cost of the highest energy consumption. MB2C ranks below PPO in the TOPSIS table, showing its dependence on model accuracy and highlighting challenges in MPC optimization. PPO ranks third overall and performs well in the January scenarios in terms of energy savings, achieving the lowest energy consumption in Chicago in January, but at the cost of higher violations. It also remains data-hungry and requires careful reward design and configuration.

Another important observation is that the LLM-based controllers perform competitively against standard control baselines in the trade-off evaluation. This suggests that LLMs can serve as practical controllers for building operation when provided with well-structured task context. Compared with conventional approaches, LLM-based control can reduce some implementation burden because it does not require manually defining a detailed optimization model, tuning MPC parameters, designing reward weights, or collecting large amounts of training data. It also provides a more interpretable control process, since the model can generate textual reasoning alongside the selected actions. However, the results also show that LLM control is highly dependent on the quality and structure of the provided context. Simple prompting or generic guidance is not sufficient to achieve the best performance.

This work does not include a real-world implementation, but the broader evaluation across diverse seasons and climates assesses ThermoLLM's performance and reliability. The framework also offers a practical advantage in portability. As the Brick-style schema is standardized and widely used in industry, ThermoLLM can be readily deployed and tested in real-world buildings using an existing Brick knowledge graph or one constructed from building information. A challenge with the current approach is that performance depends on the LLM's capacity and reasoning capabilities and requires careful prompt-template design. Cost and latency are also concerns for longer evaluations. These challenges motivate future evaluation across models of different sizes and computational costs, including small LLMs that can be deployed at the edge within building management systems. Such models could address latency, monetary cost, computational cost, and privacy concerns while maintaining performance comparable to that of large cloud-based LLMs.

Overall, the results support the central design claim of this paper: effective LLM-based decision-making in multi-zone HVAC control benefits from jointly modeling recent thermal evolution and spatial-semantic building context. History helps the model understand short-term dynamics, while KG grounding provides the structural context needed to interpret those dynamics at the zone level. The combination of both components enables ThermoLLM to achieve the strongest comfort--energy trade-off among the evaluated methods.

\section{Conclusion}

This paper presented ThermoLLM, a thermodynamics-aware LLM framework for multi-zone HVAC control that combines a Brick-derived spatial-semantic knowledge graph with recent interaction history. The main idea is that effective HVAC control depends not only on current observations and reward values but also on building structural information and thermal response. By linking spatial-semantic building context with recent operating history, ThermoLLM provides the LLM with more decision-relevant context than literature-based control strategies or static prompt guidelines. The results show that ThermoLLM ranks first in the TOPSIS table, achieving the highest relative closeness score in each scenario, and performs well against DARLIN in terms of energy--comfort performance, repeatability, and operational overhead. This suggests that the structure of the provided context is important for LLM-based control and that building-specific spatial-semantic context can be more useful than static heuristic instructions or generic literature-based retrieval. Overall, this work shows that carefully structured contextual grounding can improve the in-context decision-making ability and adaptability of LLMs, enabling them to better interpret and process complex environments.

\section*{Generative AI Disclosure}

Generative AI was used in the preparation of this manuscript in two limited ways. First, OpenAI ChatGPT (GPT-5; accessed November 2026) was used to correct grammar and refine sentence wording, including portions of the Related Work and other manuscript sections. Second, ChatGPT was used to just generate the visual rendering of Fig.~3 (the Pareto scatter plot) from numerical energy and PMV-violation values produced by our experiments. Generative AI was not used to generate or modify the underlying experimental data, analyses, citations, or scientific conclusions. All AI-assisted text and figure outputs were reviewed, corrected, and validated by the authors, who retain full responsibility for the manuscript.

% Acknowledgments and funding statement removed from anonymous review copy.
% Restore the original funding statement after acceptance.

\bibliographystyle{ACM-Reference-Format}
\bibliography{references_ubi_final}

\end{document}